\documentclass[lettersize,journal]{IEEEtran}
\usepackage{amsmath,amsfonts}
\usepackage{algorithmicx}
\usepackage{algorithm}
\usepackage{array}
\usepackage[caption=false,font=normalsize,labelfont=sf,textfont=sf]{subfig}
\usepackage{textcomp}
\usepackage{stfloats}
\usepackage{url}
\usepackage{verbatim}
\usepackage{graphicx}
\usepackage{cite}
\usepackage{mdframed}
\usepackage{listings}
\usepackage{xcolor}
\usepackage{algpseudocode}
\usepackage{tcolorbox}
\newtcolorbox{highlighted}{colback=yellow!20, colframe=yellow!40, boxrule=0pt, arc=0pt, outer arc=0pt, left=2pt, right=2pt, top=2pt, bottom=2pt}

\usepackage{tikz}
\usetikzlibrary{shapes.geometric, arrows, positioning}

\definecolor{actioninputcolor}{rgb}{1,0,1} 
\definecolor{actioncolor}{rgb}{1,0,0} 
\definecolor{observationcolor}{rgb}{0,0,1} 
\definecolor{thoughtcolor}{rgb}{0,0.6,0} 
\definecolor{finalanswercolor}{rgb}{0,0,0} 
\definecolor{codegray}{rgb}{0.5,0.5,0.5}
\definecolor{backcolour}{rgb}{0.95,0.95,0.92}

\definecolor{actioninputcolor}{rgb}{1,0,1} 
\definecolor{actioncolor}{rgb}{1,0,0} 
\definecolor{observationcolor}{rgb}{0,0,1} 
\definecolor{thoughtcolor}{rgb}{0,0.6,0} 
\definecolor{finalanswercolor}{rgb}{0,0,0} 
\definecolor{codegray}{rgb}{0.5,0.5,0.5}
\definecolor{backcolour}{rgb}{0.95,0.95,0.92}

\lstdefinestyle{pythonstyle}{
  language=Python,
  basicstyle=\ttfamily\small,
  keywordstyle=\color{blue},
  commentstyle=\color{brown},
  backgroundcolor=\color{backcolour},
  numbers=none,
  breaklines=true,
  captionpos=b,
  showspaces=false,
  showstringspaces=false,
}

\definecolor{codegreen}{rgb}{0,0.6,0}
\definecolor{codegray}{rgb}{0.5,0.5,0.5}
\definecolor{codepurple}{rgb}{0.58,0,0.82}
\definecolor{backcolour}{rgb}{0.95,0.95,0.92}

\lstdefinestyle{mystyle}{
    backgroundcolor=\color{backcolour},   
    commentstyle=\color{codegreen},
    keywordstyle=\color{magenta},
    numberstyle=\tiny\color{codegray},
    stringstyle=\color{codepurple},
    basicstyle=\ttfamily\footnotesize,
    breakatwhitespace=false,         
    breaklines=true,                 
    captionpos=b,                    
    keepspaces=true,                 
    numbers=left,                    
    numbersep=5pt,                  
    showspaces=false,                
    showstringspaces=false,
    showtabs=false,                  
    tabsize=2
}

\begin{document}

\title{SynthAI: A Multi Agent Generative AI Framework for Automated Modular HLS Design Generation}
\author{\IEEEauthorblockN{Seyed Arash Sheikholeslam}\\ \IEEEauthorblockA{International Monetary Fund (IMF)
1900 Pennsylvania Ave, Washington, DC, USA\\
ssheikholeslam@imf.org} \\
\thanks{The views expressed in this paper are those of the author(s) and do not necessarily represent the views of the IMF, its Executive Board, or IMF management.}
\and
\IEEEauthorblockN{Andre Ivanov},~\IEEEmembership{Fellow,~IEEE,}\\
\IEEEauthorblockA{UBC, Department of ECE
2332 Main Mall, Vancouver, BC, Canada\\
ivanov@ece.ubc.ca}
}

\maketitle

\begin{abstract}
In this paper, we introduce SynthAI, a new method for the automated creation of High-Level Synthesis (HLS) designs. SynthAI integrates ReAct agents, Chain-of-Thought (CoT) prompting, web search technologies, and the Retrieval-Augmented Generation (RAG) framework within a structured decision graph. This innovative approach enables the systematic decomposition of complex hardware design tasks into multiple stages and smaller, manageable modules. As a result, SynthAI produces synthesizable designs that closely adhere to user-specified design objectives and functional requirements. We further validate the capabilities of SynthAI through several case studies, highlighting its proficiency in generating complex, multi-module logic designs from a single initial prompt. The SynthAI code is provided via the following repo: 
\url{https://github.com/sarashs/FPGA_AGI}
\end{abstract}


\section{Introduction}
\IEEEPARstart{M}{ultiple} studies have bench-marked and evaluated the capabilities and limitations of Large Language Models (LLMs) in the emergent field of automated hardware code generation. We begin by providing a general overview of the field.

LLMs have shown promising capabilities in generating basic logic hardware designs. The concept of deriving hardware description code from natural language was explored by \textit{Pearce et al.} \cite{pearce2020dave}. Further advancements were discussed in the field of conversational hardware design, emphasizing the challenges and opportunities, as highlighted by \textit{Blocklove et al.} \cite{blocklove2023chip} and \textit{Chang et al.} \cite{chang2023chipgpt}. All three have shown LLMs' ability to generate basic logic code.

In the realm of specialized LLMs for code generation, \textit{Thakur et al.} \cite{thakur2023benchmarking} investigated LLMs' capacity to produce Verilog code after fine-tuning. They demonstrated that fine-tuning LLMs with Verilog datasets enhances their ability to generate syntactically correct code. Their study was groundbreaking in developing an evaluation framework for both functional analysis and syntax testing of the generated code. The main contribution of their work lies in providing open-source training, evaluation scripts, and LLM checkpoints. However, a notable limitation is their method's difficulty in managing advanced Verilog code generation tasks.

The LLM4SecHW framework, developed by \textit{Fu et al.} \cite{fu2023llm4sechw}, focuses on employing LLMs for hardware design debugging, particularly addressing the scarcity of domain-specific data. Their significant contribution was that of creating a dataset for hardware design defects, paving the way for novel applications in hardware security.

\textit{Liu et al.} \cite{liu2023verilogeval} introduced VerilogEval, a framework aimed at evaluating LLMs in Verilog code generation. Their work is distinguished by establishing the first benchmarking framework capable of conducting comprehensive evaluations across various Verilog code generation tasks.

\textit{Du et al.} \cite{du2023power} explored LLMs' application in generating HDL code for wireless signal processing, incorporating in-context learning and Chain-of-Thought (CoT) prompting. Their work stands out for its potential in generating complex HDL code for advanced applications, offering insightful contributions to wireless communication hardware.

Sequential usage of LLMs via prompt engineering can achieve only limited success, particularly when handling complex tasks in code generation. To address these challenges, multi-agent generative AI frameworks \cite{xu2024genaipoweredmultiagentparadigmsmart, zou2023wireless,islam2024mapcoder} have emerged, involving multiple AI agents that collaborate, communicate, or compete to enhance the generation process. These frameworks allow agents to specialize in distinct tasks such as reasoning, evaluation, and iterative refinement, which collectively improve the quality and coherence of the generated outputs. Multi-agent systems have been effective in areas like collaborative design, code generation, and complex reasoning, where agents provide feedback, enhance creativity, and collaborate towards solving problems.

In our work, we introduce SynthAI (Figure \ref{SynthAI-hw}), a framework that enhances large language models (LLMs) ability to generate hardware code by enabling them to use tools such as web searches and database queries in an iterative manner consisting of interactive LLM agents with specialized duties (such as data gathering, system design, coding and evaluation). This framework uses diverse prompting techniques, like Chain-of-Thought (CoT), ReAct, and in-context-learning to generate synthsizable and optimized code without retraining. SynthAI accesses real-time data and a comprehensive database with code samples and FPGA datasheets, shifting the paradigm in automated hardware code generation over previous methods, planning and executing modular designs from single prompts efficiently.
The SynthAI code is provided via the following repo: \url{https://github.com/sarashs/FPGA_AGI}

\section{LLM-based Technologies used in SynthAI}
Here, we outline the various logical inference and guided text generation techniques that serve as the foundation of our proposed framework. The core approach underlying these methods is iterative prompting. These generative processes can be generalized by the following formulation:

\begin{equation}
R_i = \text{LLM}(T_{i-1} \oplus_{j=0}^{i-1} AO_{j} \oplus \text{Anc}_{i-1}), \quad T_{i-1} \in \{T\}
\end{equation} \label{eq_1}

In this formulation:

\begin{itemize}
    \item $\{T\}$ is a set of prompt templates.
    \item $\oplus$ denotes the augmentation operation, such as concatenation or insertion.
    \item $AO_{j}$ represents the outcome of an action $A_{j} \in \{A\}$ performed outside of the LLM at stage $j$. The cumulative augmentation operation means that we are keeping the action outcomes from the previous steps of a generative process.
    \item $\text{Anc}_i$ is ancillary text inserted into the prompt by the system overseeing the LLM's operation.
    \item $\text{LLM}$ denotes the LLM's generation process at iteration $i$, which produces the response $R_i$.
\end{itemize}

This approach systematically incorporates external actions and additional contextual information, enhancing the LLM's output through guided, iterative prompting.
\subsection{MRKL}

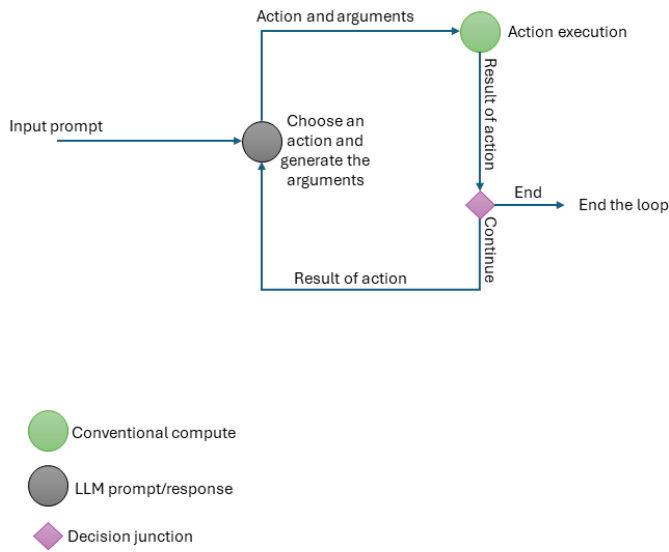
\begin{figure}[!t]
\centering
\begin{tikzpicture}[node distance=1cm, font=\sffamily, scale=0.75, transform shape] 

\tikzstyle{process} = [circle, draw, text width=1.8cm, minimum size=1.8cm, align=center]
\tikzstyle{action} = [circle, draw, text width=1.8cm, minimum size=1.8cm, align=center]
\tikzstyle{decision} = [diamond, draw, text centered, aspect=2, minimum width=2cm, inner sep=0pt]
\tikzstyle{arrow} = [thick,->,>=stealth]

\node (input) [process] {Input};
\node (action_gen) [process, right=1.5cm of input] {Action Generation Node};
\node (execution) [action, above=1.5cm of action_gen] {Action execution Node};
\node (decision) [decision, below right=0.75cm and 1.5cm of execution] {Decision Node};

\draw[arrow] (input) -- (action_gen);
\draw[arrow] (action_gen) -- node[left, text width=1.8cm] {Action and arguments} (execution);
\draw[arrow] (execution.east) -| node[above right] {Result of action} (decision.north);
\draw[arrow] (decision) -- node[below] {End} ++(2.5cm,0) node[right, text width=1.8cm] {End the loop};
\draw[arrow] (decision.south) |- node[below right] {Continue} (action_gen.east);

\end{tikzpicture}
\caption{Individual agent. Each of the blue nodes in figures \ref{SynthAI} and \ref{SynthAI-hw} is an agent similar to the one depicted here.}
\label{single_node}
\end{figure}

MRKL (Modular Reasoning, Knowledge, and Language) enhances LLMs by integrating external data, such as HLS C++ code or digital design guides, and enabling autonomous Python execution \cite{karpas2022mrkl}. ReAct agents within MRKL establish reasoning-action loops, enhancing LLM decision-making through iterative thought, action, and observation \cite{yao2022react}.
Equation \eqref{eq_1} encapsulates LLM-guided systems like Chain-of-Thought, RAG, and ReAct agents by defining input augmentation in prompts. For instance, executing Python code incorporates results via $AO_{i-1}$, with optional ancillary inputs $\text{Anc}_{i-1}$. Figure \ref{single_node} illustrates our implementation of ReAct agent, showcasing its iterative reasoning and action framework. Note that all of the agents we developed are structurally equivalent but their set of prompt templates $\{T\}$, actions $\{A\}$ and control procedures $\{Anc\}$ are different. The design agent given below is just an example. Let's take a step-by-step look into that example.

Here is an example of the part of the prompt template for our module design agent:

\begin{mdframed}
\begin{lstlisting}[style=pythonstyle]
"""
You are an FPGA design engineer responsible for writing synthesizable code for an HLS hardware project. 

\\The guidelines for the module designer were removed by the authors to save space

Thought: You should think of an action. You do this by calling the Thought tool/function. This is the only way to think.
Action: You take an action through calling one of the search_web or python_run tools.
... (this Thought/Action can repeat 3 times)
Response: You Must use the CodeModuleResponse tool to format your response. Do not return your final response without using the CodeModuleResponse tool
"""
\end{lstlisting}
\end{mdframed}
Algorithm \ref{alg:agent_operation} showcases our module design agent. Note that we have incorporated \textit{Chain of Thought} (CoT) reasoning \cite{wei2022chain, zhang2022automatic} into the agent's reasoning process. Chain of Thought (CoT) prompting is a method that enhances the reasoning capabilities of LLMs by forcing them to iteratively generate their reasoning before arriving at a conclusion or making a decision. Our innovative approach fuses the ReAct framework with CoT by treating "thought" as a distinct action within the agent's operational cycle. This synthesis allows the agent to perform an arbitrary number of thought iterations, facilitating deeper reasoning, before committing to a final action. By combining these methodologies, our agent achieves enhanced flexibility and robustness in its reasoning processes, leading to more informed and accurate decision-making. 

\begin{algorithm}
\caption{Module generation Agent with Chain-of-Thought reasoning}
\label{alg:agent_operation}
\begin{algorithmic}[1]
\State \textbf{Input:} Prompt template set $\{T\}$
\State \textbf{Output:} Final hardware design code
\State Initialize $i \gets 0$
\While{design task not complete}
    \State \textbf{Thought Action:} Generate reasoning step
    \State $T_i \gets$ Update template from $\{T\}$
    \State $R \gets \text{LLM}(T_{i-1} \oplus_{j=0}^{i-1} AO_{j} \oplus \text{Anc}_{i-1})$ 
    \Comment{Reasoning output}
    
    \State \textbf{Evaluate Thought:} Determine next action based on $R$
    
    \If{decision is Web Search}
        \State \textbf{Action:} Perform web search
        \State $AO_{i} \gets \text{Execute}(\text{Web Search})$
    \ElsIf{decision is Python Code Execution}
        \State \textbf{Action:} Execute Python code
        \State $AO_{i} \gets \text{Execute}(\text{Python Code})$
    \Else
        \State \textbf{Thought Action:} Continue reasoning
        \State $AO_{i} \gets \text{LLM}(T_i \oplus_{j=0}^{i-1} AO_{j} \oplus R \oplus \text{Anc}_i)$
    \EndIf
    
    \State Increment $i$
\EndWhile

\State \textbf{Final Action:} Format the result into hardware design code
\State \textbf{Output} $\gets \text{CodeModuleResponse}(AO_i)$

\end{algorithmic}
\end{algorithm}

\subsection{Interactive Agents}

\begin{figure*}[!t]
\centering
\begin{tikzpicture}[node distance=2cm, font=\sffamily, scale=0.55, transform shape]

\tikzstyle{blue_node} = [circle, draw=blue, fill=blue!20, text width=1.8cm, minimum size=1.8cm, align=center]
\tikzstyle{circle_node} = [circle, draw, text width=1.8cm, minimum size=1.8cm, align=center]
\tikzstyle{normal_node} = [rectangle, draw, text centered, minimum height=2.5cm, minimum width=3cm]
\tikzstyle{DB} = [cylinder, cylinder uses custom fill, shape border rotate=90, aspect=0.25, draw, minimum height=2cm, minimum width=1cm, text width=1.8cm, text centered]
\tikzstyle{PY} = [rectangle, draw, text centered, minimum height=2cm, text width=2cm, minimum width=1.8cm]
\tikzstyle{decision} = [diamond, draw, text centered, aspect=2, minimum width=2cm, inner sep=0pt]
\tikzstyle{arrow} = [thick,->,>=stealth]

\node (objectives) [align= center, above= 1.5cm of input] {Objectives};
\node (lit_review) [align= center, left= 1.5cm of input] {Literature review};
\node (input) [circle_node] {Input Prompt};
\node (sys) [blue_node, right=2cm of input] {System Design};
\node (sys_eval) [blue_node, right=2cm of sys] {Design evaluation};
\node (sys_decision) [decision, right=2cm of sys_eval] {Decision node};
\node (redesign) [blue_node, above=1.5cm of sys_decision] {Redesign};
\node (sort) [circle_node, right =2cm of sys_decision] {Topological sort};
\node (single_mod) [blue_node, below =2cm of sort] {Module design};
\node (mod_integrate) [blue_node, left =2cm of single_mod] {Module integration};
\node (final_eval) [blue_node, left =2cm of mod_integrate] {Final design evaluation};
\node (final_decision) [decision, left=2cm of final_eval] {Decision node};
\node (files) [circle_node, left=2cm of final_decision] {Generate files};

\node (project_db) [DB, below =2cm of single_mod, align=center] {Language specific manuals};
\node (generic_kb) [PY, below=2cm of mod_integrate, align=center] {Computation (python shell)};

\draw[arrow] (objectives) -- (input);
\draw[arrow] (lit_review) -- (input);
\draw[arrow] (input) -- (sys);
\draw[arrow] (sys) -- (sys_eval);
\draw[arrow] (sys_eval) -- (sys_decision);
\draw[arrow] (sys_decision.north) -- (redesign.south);
\draw[arrow] (sys_decision.east) -- (sort);
\draw[arrow] (redesign.west) -| (sys_eval.north);
\draw[arrow] (sort) -- (single_mod);
\draw[arrow] (single_mod) -- (mod_integrate);
\draw[arrow] (mod_integrate) -- (final_eval);
\draw[arrow] (final_eval) -- (final_decision.east);
\draw[arrow] (final_decision.west) -- (files);
\draw[arrow] (final_decision.north) |- ++(0,1) -| (mod_integrate.north);

\draw[arrow] (project_db) -- (single_mod);
\draw[arrow] (generic_kb.north) |- ++(0,1) -| (single_mod.south);
\draw[arrow] (generic_kb.north) -- (mod_integrate);

\end{tikzpicture}
\caption{Flow diagram of hardware design component consisting of various agents and evaluators.}
\label{SynthAI-hw}
\end{figure*}
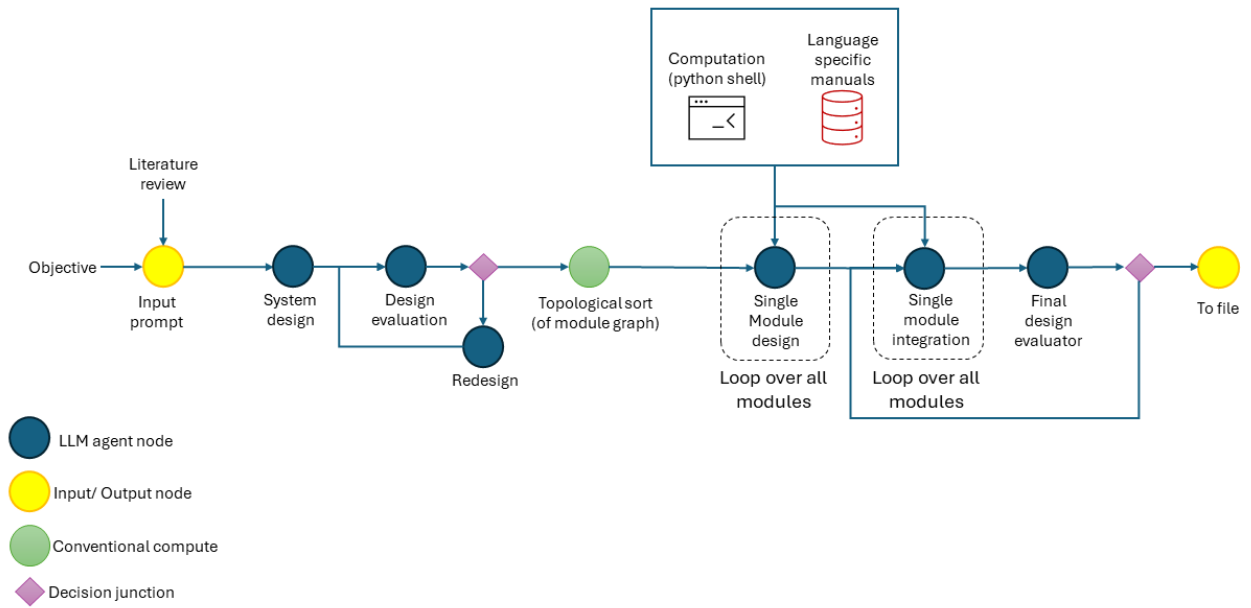

The latest LLM-based technology that we incorporate is that of Interactive Agents. The idea is that task specific agents (similar to above) will work within a team to perform, planning, task break down and task execution \cite{shi2024learning, guo2024embodied, abdelnabi2023llm}. In the case of our work, we have data collection agents, system design agents, module design agents, and evaluation agents, and these interact with each other according to the on the graph depicted in Figure \ref{SynthAI-hw}.

\subsection{Retrieval Augmented Generation (RAG)}
Another technique we use in SynthAI is Retrieval Augmented Generation (RAG) \cite{gao2023retrieval, levonian2023retrieval, ding2023realgen}, a  method that augments LLMs knowledge by incorporating external knowledge retrieval into their generative process. In practice, RAG involves constructing a vector database from semantic data. This database acts as a search tool for the LLM, enabling it to supplement its inherent knowledge (embedded in its parameters) with external, task-specific information. RAG allows us to integrate diverse resources like textbooks and coding best practices and other domain-specific knowledge.

\begin{figure*}[!t]
\centering
\begin{tikzpicture}[node distance=2cm, font=\sffamily, scale=0.6, transform shape]

\tikzstyle{blue_node} = [circle, draw=blue, fill=blue!20, text width=1.8cm, minimum size=1.8cm, align=center]
\tikzstyle{circle_node} = [circle, draw, text width=1.8cm, minimum size=1.8cm, align=center]
\tikzstyle{normal_node} = [rectangle, draw, text centered, minimum height=2.5cm, minimum width=3cm]
\tikzstyle{DB} = [cylinder, cylinder uses custom fill, shape border rotate=90, 
                  aspect=0.25, draw, minimum height=2cm, minimum width=1cm, text width=1.8cm, text centered]
\tikzstyle{decision} = [diamond, draw, text centered, aspect=2, minimum width=2cm, inner sep=0pt]
\tikzstyle{arrow} = [thick,->,>=stealth]

\node (objectives) [align= center, above= 1.5cm of input] {Objectives};
\node (input) [circle_node] {Input Prompt};
\node (gen_questions) [blue_node, right=2cm of input] {Generate questions};
\node (rag) [blue_node, right=2cm of gen_questions] {RAG};
\node (evaluator) [blue_node, right=2cm of rag] {Evaluator};
\node (decision) [decision, right=2cm of evaluator] {Decision Node};
\node (web_search) [circle_node, above =1.5cm of decision] {Web search};
\node (lit_review) [blue_node, below =1.5cm of decision] {Review Generation};
\node (final_output) [circle_node, left=2cm of lit_review] {Literature review};

\node (project_db) [DB, above left=2.5cm and 1cm of rag, align=center] {Project specific database};
\node (generic_kb) [DB, right=1.5cm of project_db, align=center] {Generic Knowledge-base};

\draw[arrow] (objectives) -- (input);
\draw[arrow] (input) -- (gen_questions);
\draw[arrow] (gen_questions) -- (rag);
\draw[arrow] (rag) -- (evaluator);
\draw[arrow] (evaluator) -- (decision);
\draw[arrow] (decision.north) -- (web_search);
\draw[arrow] (decision.south) -- (lit_review);
\draw[arrow] (web_search.west) -| (evaluator.north);
\draw[arrow] (lit_review) -- (final_output);

\draw[arrow] (project_db) -- (rag);
\draw[arrow] (generic_kb) -- (rag);

\end{tikzpicture}
\caption{Flow diagram of knowledge gathering component consisting of various agents, evaluators, and a vector database. The blue nodes are agents that employ LLMs and have an architecture similar to Figure~\ref{single_node}}
\label{SynthAI}
\end{figure*}

\section{SynthAI Architecture}

Our project focuses on developing a technology demonstrator and proof of concept that converts user-defined design objectives into comprehensive hardware solutions. At the heart of our method is a series of specialized agents within the SynthAI system. These agents collaboratively interpret the initial objectives, refine them into detailed designs, and ultimately generate synthesizable HLS code, complete with corresponding test benches.

The SynthAI system is divided into two main components: a knowledge gathering component, which functions similarly to an automated literature review tool, and a hardware design component. These are depicted in Figures \ref{SynthAI} and \ref{SynthAI-hw}, respectively. Each component comprises individually tasked agents, each with a specific role. These agents operate within a decision graph that determines subsequent steps based on the current state and the results of each agent's task execution. This process mimics a state machine, ensuring that tasks are completed sequentially and effectively, moving from one state to the next based on predefined rules and the outcomes of actions taken at each stage.

The knowledge gathering component primarily involves a Retrieval-Augmented Generation (RAG) system that processes various engineering documents and past designs, enhanced by project-specific texts and internet searches. A question-generating agent formulates queries reflecting the project's goals and specific requirements. These queries are initially matched against a local database. If the database yields insufficient information, an internet search is conducted to enrich the response further. The resulting data, along with project requirements and specific texts, are then synthesized by another agent into a design-focused literature review. We have intentionally kept this description concise for brevity. For those interested in the detailed workings of this component, we invite you to consult the code available in our repository. The algorithm governing the multi-agent process of data gathering, evaluation and websearch is given in the following.

\begin{algorithm}
\caption{Retrieval Augmented Generation with Evaluation Loop}
\label{alg:rag_evaluation}
\begin{algorithmic}[1]
\Require $\text{LLM1}$: Primary language model, $\text{LLM2}$: Evaluation language model, $\mathcal{D}$: Set of external resources (e.g., textbooks, coding guides)
\State \textbf{Initialize} $\text{VectorDB} \leftarrow \text{ConstructVectorDatabase}(\mathcal{D})$ \Comment{Build a vector database from external resources}
\State \textbf{Initialize} $T \leftarrow$ Select prompt template from $\{T\}$
\State $\text{Anc} \leftarrow \text{InitializeAncillaryText()}$ \Comment{Initialize ancillary text if required}
\State $\text{Satisfied} \leftarrow \text{False}$
\State Initialize $i \gets 0$
\State $Q \leftarrow \text{LLM1}(T \oplus_{j=0}^{i-1} AO_{j} \oplus \text{Anc}_{i-1})$ \Comment{Generate query based on quary generation prompt template T}
\State $AO_{i} \leftarrow \text{Retrieve}(\text{VectorDB}, Q)$ \Comment{Retrieve external knowledge relevant to the query}
\State $R_i \leftarrow \text{LLM1}(T^G \oplus AO_{i} \oplus \text{Anc}_{i-1})$ \Comment{Generate response using response generation prompt $T^G$}
\While{not $\text{Satisfied}$}
    \State $\text{Evaluation} \leftarrow \text{LLM2}(T^E \oplus_{j=0}^{i-1} AO_{j} \oplus R_i)$ \Comment{LLM2 evaluates the generated response based on prompt template $T^E$}
    \If{$\text{Evaluation} = \text{Satisfactory}$}
        \State $\text{Satisfied} \leftarrow \text{True}$
    \Else
        \State $Q_{\text{search}} \leftarrow \text{GenerateSearchQuery}(\text{LLM2}, R_i)$ \Comment{Generate a search query based on the evaluation}
        \State $AO_{i} \leftarrow \text{InternetSearch}(Q_{\text{search}})$
        \Comment{Perform internet search and update $AO_i$}
        \State $R_{i} \leftarrow AO_{i}$ 
    \EndIf
\EndWhile
\State $R_{\text{final}} \leftarrow \text{LLM2}(T^R \oplus R_i)$ \Comment{LLM2 generates the final literature review}
\Ensure $R_{\text{final}}$: Final literature review generated after evaluation
\end{algorithmic}
\end{algorithm}

\subsection {Hardware design}

In the hardware design component of our system, the initial input consists of design objectives—both goals and requirements—supplemented by the outcomes of the literature review. The first agent in this sequence undertakes the system-level design, which is crucial for setting the foundational architecture of the hardware. This initial design is then assessed by an evaluation agent, which scrutinizes the system design and provides essential feedback for any necessary redesigns. This iterative process ensures that the design is both efficient and aligns with the specified objectives.

The refined system design is articulated through a list of modules, formatted according to the specifications outlined in the Python code snippet below. This format serves as a directive for the LLM, guiding the structured development of HLS code by detailing each module's name, functionality, connections, and ports, along with a templated outline for the code to be completed by a coder. This structured approach facilitates a clear, organized path from design conception to ready-to-implement hardware modules.

The list of modules, or the system graph, undergoes a topological sorting process before being passed to a module coder. This sorting ensures that modules with fewer connections are coded first, enhancing interface consistency across the system. The module designer agent is equipped with capabilities for coding syntax searches, and potentially, internet searches, as well as access to a Python shell. These tools enable the agent to research unfamiliar coding practices and perform computations beyond the conventional capabilities of large language models (LLMs). For instance, the agent might use the Python shell to calculate twiddle factors for a Fast Fourier Transform (FFT), which will be detailed in the subsequent section. 
Following the preliminary module design, the modules are processed by a module integrator. This stage involves replacing any remaining placeholders or simplified code with fully synthesizable and optimized code. Finally, a comprehensive evaluation is conducted by a final evaluator, who assesses the coding practices, completeness, synthesizability, and optimization of the modules in alignment with the design requirements. The evaluator continues to provide feedback to the integration agent until the integrated design meets all criteria and is approved for saving into files. This iterative refinement ensures that the final design not only meets the initial objectives but also adheres to the highest standards of quality and efficiency.

\section{Case studies}

Our work builds on the foundations laid by \textit{Thakur et al.} \cite{thakur2023benchmarking} and \textit{Du et al.} \cite{du2023power}. However, due to the rapid advancements in LLM-based technologies, a direct comparison between our work and the aforementioned studies would be impractical. Instead, to assess the performance of SynthAI, we present three circuit design case studies and evaluate its performance using the following metrics:

\begin{itemize}
    \item \textbf{System Design}: This metric evaluates whether the modules' graph and their connections are logically sound and adhere to design specifications. For example, in a CPU design, we expect the presence of ALU, datapath, memory, and control modules or their equivalents in the graph.
    
    \item \textbf{Syntax}: This metric assesses if the generated code is syntactically correct and follows AMD HLS C++ guidelines.
    
    \item \textbf{Completeness}: This metric measures whether the generated design fully implements the intended functionality without missing any critical components or features. A design that contains superfluous code or placeholders instead of actual code is considered incomplete.
    
    \item \textbf{Interface}: This metric evaluates the correctness of the interfaces between modules, ensuring that they match in width, type, and integration within the system.
    
    \item \textbf{Optimization}: This metric analyzes the efficiency of the generated design in terms of performance metrics per design goals and requirements. In HLS, some optimizations are performed via compiler pragmas, such as unrolling loops or partitioning memory.
    
    \item \textbf{Synthesizability}: This metric assesses whether the generated design can be successfully synthesized into hardware, verifying that it meets all necessary constraints and requirements for physical implementation. Although we do not perform extensive tests for this step, we rely on error-free synthesizability and the passing of basic functionality tests.
\end{itemize}

For our three circuit designs we have the following:
\subsection*{Fast Fourier Transform:}
Goals:
\begin{itemize}
   \item Design a 128-point FFT circuit using two pre-existing 64-point FFT modules.
   \item Optimize the design for maximum performance, specifically focusing on speed.
   \item Ensure the design is implemented in HLS C++ and is well-commented, particularly highlighting performance optimizations.
\end{itemize}
Requirements: 
\begin{itemize}
   \item The circuit must accept an input array of 128 double precision fixed point real numbers.
   \item Utilize the radix-2 DIT FFT algorithm to compute the 128-point FFT by leveraging two 64-point FFT modules.
   \item Implement the design in HLS C++ with detailed comments explaining performance optimizations.
   \item The design should efficiently handle the decomposition of the 128-point FFT into two 64-point FFTs, processing even-indexed and odd-indexed inputs separately.
   \item After computing the 64-point FFTs, the results should be combined using the radix-2 DIT method to produce the final 128-point FFT output.
   \item Ensure that the design adheres to the mathematical principles of the radix-2 DIT FFT, including the correct handling of the complex exponential factors and the summation processes.
\end{itemize}

\subsection*{quadruple precision floating point exponentiation:}
Goals:
\begin{itemize}
   \item Design and implement a quadruple precision floating point exponentiation module using a 128-bit floating point representation.
   \item Ensure the module adheres to the IEEE 754-2008 binary128 standard for quadruple precision floating point numbers.
   \item Optimize the module for accurate and reliable computation, minimizing overflow and round-off errors.
\end{itemize}
Requirements:
\begin{itemize}
   \item Implement a 128-bit floating point representation with 1 sign bit, 15 exponent bits, and 113 significand bits (112 explicitly stored, 1 implicit).
   \item Follow the IEEE 754-2008 standard for binary128, including the encoding of the exponent using an offset binary representation with a bias of 16383.
   \item Handle special cases in the floating point representation such as zero, subnormal numbers, infinity, and NaN (Not a Number).
   \item Implement exponentiation operations that can handle the full range of quadruple precision values, from the smallest subnormal to the largest normal number.
   \item Ensure that the conversion between decimal and binary128 representations is accurate, adhering to the precision requirements specified in the IEEE standard.
   \item Develop test cases covering typical, boundary, and special values to verify the correctness and precision of the floating point operations and exponentiation logic.
\end{itemize}

\subsection*{simple educational RISC-V CPU:}
Goals:
\begin{itemize}
   \item Design a simple educational RISC-V CPU using HLS C++.
   \item Ensure the HLS C++ code is synthesizable and fits on a Zynq-7 FPGA device.
   \item Provide extensive comments within the code for educational purposes.
\end{itemize}
Requirements:
\begin{itemize}
   \item Implement a RISC-V CPU architecture capable of executing basic RISC-V instructions such as load, store, add, subtract, logical operations, and control flow.
   \item Design the CPU to be compatible with the Zynq-7 FPGA, ensuring efficient use of resources to fit the design within the FPGA constraints.
   \item Utilize HLS C++ for the implementation, leveraging its features to efficiently map high-level constructs to hardware.
   \item Include detailed comments explaining each part of the code and its functionality, suitable for educational purposes.
   \item Ensure the design includes a memory interface, ALU, register file, and control unit, each described clearly in HLS C++.
   \item Implement test benches in HLS C++ to verify each component of the CPU, including memory operations, arithmetic operations, and control flow logic.
\end{itemize}

Table \ref{comparison_table} presents the results of our case studies, comparing the performance of GPT-3.5-Turbo, GPT-4-Turbo, and GPT-4o (the latest model at the time of writing). We used separate LLMs for the nodes responsible for evaluating other nodes. Although the specific use of different LLMs is not depicted, Figure \ref{SynthAI-hw} illustrates the system design and includes the various nodes of the system. This approach was partly to determine if a proficient evaluator could offset the lower performance of a weaker (and more cost-effective) model used for system design and module coding.

In our initial evaluations, we observed that GPT-3.5-Turbo was not an effective evaluator, as it consistently approved the system design and system code across all metrics. Consequently, we decided to use only GPT-4-Turbo and GPT-4o as evaluators.

For each combination of [design problem, design model, evaluator], we conducted five runs using SynthAI, selecting the optimal design from these trials. The results are summarized in Table \ref{comparison_table}.  Most experiments encountered failures, primarily due to syntax errors, synthesizability issues, or systemic design flaws. Nonetheless, we consistently identified one or two promising outcomes per test set. Initial observations reveal that while LLMs can integrate optimization pragmas to enhance performance (via pipe-lining or loop unrolling) and reduce resource usage, their implementation is not consistently technically sound or efficient. Furthermore, the system lacks the capability to optimize based on post-synthesis performance metrics, due to the absence of such feedback by design.

\definecolor{darkgreen}{rgb}{0.0, 0.5, 0.0}
\newcommand{\sharedfootnoteone}{\footnote{Some variables were used but were not declared in the top module.}}
\begin{table*}[h!]
\centering
\begin{tabular}{|c|c|c|c|c|c|c|}
\hline
\textbf{Evaluator} & \multicolumn{3}{c|}{\textbf{GPT-4o}} & \multicolumn{3}{c|}{\textbf{GPT-4-Turbo}}  \\ \hline
\textbf{Design Model} & GPT-3.5-Turbo & GPT-4-Turbo & GPT-4o & GPT-3.5-Turbo & GPT-4-Turbo & GPT-4o \\ \hline
& \multicolumn{6}{c|}{\textbf{Simple Risc-V}}  \\ \hline
\textbf{system design} & \textcolor{darkgreen}{Pass} & \textcolor{darkgreen}{Pass} & \textcolor{darkgreen}{Pass} & \textcolor{red}{Fail} & \textcolor{darkgreen}{Pass} & \textcolor{darkgreen}{Pass} \\ \hline
\textbf{Syntax} & \textcolor{darkgreen}{Pass} & \textcolor{darkgreen}{Pass} & \textcolor{darkgreen}{Pass} & \textcolor{red}{Fail} & \textcolor{darkgreen}{Pass} & \textcolor{darkgreen}{Pass} \\ \hline
\textbf{interface} & \textcolor{red}{Fail} & \textcolor{darkgreen}{Pass} & \textcolor{red}{Fail}$^c$ & \textcolor{red}{Fail} & \textcolor{darkgreen}{Pass} &  \textcolor{darkgreen}{Pass} \\ \hline
\textbf{Completeness} & \textcolor{red}{Fail} & \textcolor{darkgreen}{Pass} & \textcolor{darkgreen}{Pass} & \textcolor{red}{Fail} & \textcolor{red}{Fail} $^a$ &  \textcolor{darkgreen}{Pass}\\ \hline
\textbf{Optimization} & \textcolor{red}{Fail} & \textcolor{darkgreen}{Pass} $^b$ & \textcolor{darkgreen}{Pass} & \textcolor{red}{Fail} & \textcolor{darkgreen}{Pass} $^b$ & \textcolor{darkgreen}{Pass} $^b$\\ \hline
\textbf{synthesizabe} & \textcolor{red}{Fail} & \textcolor{darkgreen}{Pass} & \textcolor{darkgreen}{Pass} $^d$ & \textcolor{red}{Fail} & \textcolor{red}{Fail} & \textcolor{darkgreen}{Pass}\\ \hline
 & \multicolumn{6}{c|}{\textbf{128 point FFT}}  \\ \hline
\textbf{system design} & \textcolor{darkgreen}{Pass} & \textcolor{darkgreen}{Pass} & \textcolor{darkgreen}{Pass} & \textcolor{red}{Fail} & \textcolor{darkgreen}{Pass} & \textcolor{darkgreen}{Pass} \\ \hline
\textbf{Syntax} & \textcolor{darkgreen}{Pass} & \textcolor{darkgreen}{Pass} & \textcolor{darkgreen}{Pass} & \textcolor{red}{Fail} & \textcolor{darkgreen}{Pass} & \textcolor{darkgreen}{Pass} \\ \hline
\textbf{interface} & \textcolor{red}{Fail} & \textcolor{darkgreen}{Pass} & \textcolor{red}{Fail}$^c$ & \textcolor{red}{Fail} & \textcolor{darkgreen}{Pass} & \textcolor{darkgreen}{Pass}  \\ \hline
\textbf{Optimization} & \textcolor{red}{Fail} & \textcolor{darkgreen}{Pass} & \textcolor{darkgreen}{Pass} & \textcolor{red}{Fail} & \textcolor{red}{Fail} & \textcolor{darkgreen}{Pass} \\ \hline
\textbf{Completeness} & \textcolor{red}{Fail} & \textcolor{darkgreen}{Pass} & \textcolor{darkgreen}{Pass} & \textcolor{red}{Fail} & \textcolor{red}{Fail} & \textcolor{darkgreen}{Pass}$^e$ \\ \hline
\textbf{synthesizabe} & \textcolor{red}{Fail} & \textcolor{darkgreen}{Pass} & \textcolor{red}{Fail} & \textcolor{red}{Fail} & \textcolor{red}{Fail} & \textcolor{darkgreen}{Pass} \\ \hline
 & \multicolumn{6}{c|}{\textbf{Quadruple precision float}}  \\ \hline
\textbf{system design} & \textcolor{red}{Fail} & \textcolor{darkgreen}{Pass} & \textcolor{darkgreen}{Pass} & \textcolor{darkgreen}{Pass} & \textcolor{darkgreen}{Pass} & \textcolor{darkgreen}{Pass} \\ \hline
\textbf{Syntax} & \textcolor{darkgreen}{Pass} & \textcolor{red}{Fail}$^g$ & \textcolor{darkgreen}{Pass} & \textcolor{darkgreen}{Pass} & \textcolor{darkgreen}{Pass} & \textcolor{darkgreen}{Pass} \\ \hline
\textbf{interface} & \textcolor{red}{Fail}  & \textcolor{red}{Fail} & \textcolor{darkgreen}{Pass} & \textcolor{red}{Fail} & \textcolor{red}{Fail}$^f$ & \textcolor{darkgreen}{Pass} \\ \hline
\textbf{Optimization} & \textcolor{red}{Fail} & \textcolor{red}{Fail} &  \textcolor{darkgreen}{Pass} & \textcolor{red}{Fail} & \textcolor{red}{Fail} & \textcolor{darkgreen}{Pass} \\ \hline
\textbf{Completeness} & \textcolor{red}{Fail} & \textcolor{red}{Fail} & \textcolor{darkgreen}{Pass} & \textcolor{red}{Fail} & \textcolor{red}{Fail} & \textcolor{red}{Fail}$^h$ \\ \hline
\textbf{synthesizabe} & \textcolor{red}{Fail} & \textcolor{red}{Fail} & \textcolor{darkgreen}{Pass} & \textcolor{red}{Fail} & \textcolor{red}{Fail} & \textcolor{darkgreen}{Pass} \\ \hline
\end{tabular}\label{comparison_table}
\caption{Hierarchical evaluation of models across different experiments. The system design is evaluated by the authors while the other items in the table were based on the error/warnings produced by the EDA tool.\\\footnotesize{$^a$ Some signals were used but were not declared in the top module}\\\footnotesize{$^b$ Pragmas were correctly used to pipeline and unroll the loops and to partition the memory}\\\footnotesize{$^c$ Superfluous and meaningless directives were used.}\\\footnotesize{$^d$ Pass after removing superfluous interface directives.}\\\footnotesize{$^e$ Some methods were defines more than once and there were missing includes but after such minor changes, the code was complete.}\\\footnotesize{$^f$ Interface type conversion between input/output signals were not properly implemented.}\\\footnotesize{$^g$ The Syntax was C++ but not HLS C++ syntax.}\\\footnotesize{$^h$ While the design is complete, it is a simplified design.}}
\end{table*}

Table \ref{comparison_table} demonstrates that the best performance is achieved with GPT-4o. This indicates significant progress in large language models (LLMs), suggesting their potential for application in decision graphs, such as the one depicted in Figure \ref{SynthAI-hw}. Note that while  These models show some potential in hierarchical system design, module design, and coding.

\section{Conclusion}
The SynthAI framework we developed and described here is a multi-agent system with document search and Python execution capabilities. SynthAI represents a significant step forward in automated high-level synthesis (HLS) generation, particularly for complex electronic design tasks. While SynthAI outperforms traditional LLM-based methods in accuracy and modular structuring, it still struggles with handling complex algorithms without detailed instructions. This highlights the ongoing need for human collaboration in AI-driven processes.
 Despite these challenges, SynthAI represents an early endeavor to establish a comprehensive framework for end-to-end hardware design. The system's partial adaptability to straightforward design goals and requirements indicates significant potential for future advancements in electronic design automation.


\vfill


\begin{thebibliography}{1}

\bibitem{shi2024learning}
Z. Shi, S. Gao, X. Chen, L. Yan, H. Shi, D. Yin, Z. Chen, P. Ren, S. Verberne, and Z. Ren, 
"Learning to Use Tools via Cooperative and Interactive Agents," 
\textit{arXiv preprint arXiv:2403.03031}, 2024.

\bibitem{guo2024embodied}
X. Guo, K. Huang, J. Liu, W. Fan, N. Vélez, Q. Wu, H. Wang, T. L. Griffiths, and M. Wang,
"Embodied LLM Agents Learn to Cooperate in Organized Teams," 
\textit{arXiv preprint arXiv:2403.12482}, 2024.

\bibitem{abdelnabi2023llm}
S. Abdelnabi, A. Gomaa, S. Sivaprasad, L. Schönherr, and M. Fritz, 
"LLM-Deliberation: Evaluating LLMs with Interactive Multi-Agent Negotiation Games," 
\textit{arXiv preprint arXiv:2309.17234}, 2023.

\bibitem{wang2022self}
X. Wang, J. Wei, D. Schuurmans, Q. Le, E. Chi, S. Narang, A. Chowdhery, and D. Zhou, 
"Self-Consistency Improves Chain of Thought Reasoning in Language Models," 
\textit{arXiv preprint arXiv:2203.11171}, 2022.

\bibitem{carbonell1998use}
J. Carbonell and J. Goldstein, 
"The Use of MMR, Diversity-Based Reranking for Reordering Documents and Producing Summaries," 
in \textit{Proceedings of the 21st Annual International ACM SIGIR Conference on Research and Development in Information Retrieval}, 1998, pp. 335--336.

\bibitem{pearce2020dave}
H. Pearce, B. Tan, and R. Karri, 
"Dave: Deriving Automatically Verilog from English," 
in \textit{Proceedings of the 2020 ACM/IEEE Workshop on Machine Learning for CAD}, 2020, pp. 27--32.

\bibitem{blocklove2023chip}
J. Blocklove, S. Garg, R. Karri, and H. Pearce, 
"Chip-Chat: Challenges and Opportunities in Conversational Hardware Design," 
\textit{arXiv preprint arXiv:2305.13243}, 2023.

\bibitem{chang2023chipgpt}
K. Chang, Y. Wang, H. Ren, M. Wang, S. Liang, Y. Han, H. Li, and X. Li, 
"ChipGPT: How Far Are We from Natural Language Hardware Design?" 
\textit{arXiv preprint arXiv:2305.14019}, 2023.

\bibitem{wei2022chain}
J. Wei, X. Wang, D. Schuurmans, M. Bosma, F. Xia, E. Chi, Q. Le, and D. Zhou, 
"Chain-of-Thought Prompting Elicits Reasoning in Large Language Models," 
\textit{Advances in Neural Information Processing Systems}, vol. 35, pp. 24824--24837, 2022.

\bibitem{zhang2022automatic}
Z. Zhang, A. Zhang, M. Li, and A. Smola, 
"Automatic Chain of Thought Prompting in Large Language Models," 
\textit{arXiv preprint arXiv:2210.03493}, 2022.

\bibitem{ding2023realgen}
W. Ding, Y. Cao, D. Zhao, C. Xiao, and M. Pavone, 
"RealGen: Retrieval Augmented Generation for Controllable Traffic Scenarios," 
\textit{arXiv preprint arXiv:2312.13303}, 2023.

\bibitem{levonian2023retrieval}
Z. Levonian, C. Li, W. Zhu, A. Gade, M.-E. Postle, and W. Xing, 
"Retrieval-Augmented Generation to Improve Math Question-Answering: Trade-offs Between Groundedness and Human Preference," 
\textit{arXiv preprint arXiv:2310.03184}, 2023.

\bibitem{gao2023retrieval}
Y. Gao, Y. Xiong, X. Gao, K. Jia, J. Pan, Y. Bi, Y. Dai, J. Sun, and H. Wang, 
"Retrieval-Augmented Generation for Large Language Models: A Survey," 
\textit{arXiv preprint arXiv:2312.10997}, 2023.

\bibitem{langchain2022}
H. Chase, 
"LangChain," 2022. [Online]. Available: \url{https://github.com/langchain-ai/langchain}. [Accessed: your access date].

\bibitem{karpas2022mrkl}
E. Karpas, O. Abend, Y. Belinkov, B. Lenz, O. Lieber, N. Ratner, Y. Shoham, H. Bata, Y. Levine, and K. Leyton-Brown, 
"MRKL Systems: A Modular, Neuro-Symbolic Architecture That Combines Large Language Models, External Knowledge Sources and Discrete Reasoning," 
\textit{arXiv preprint arXiv:2205.00445}, 2022.

\bibitem{harris2021digital}
S. Harris and D. Harris, 
\textit{Digital Design and Computer Architecture, RISC-V Edition}, 
Morgan Kaufmann, 2021.

\bibitem{yao2022react}
S. Yao, J. Zhao, D. Yu, N. Du, I. Shafran, K. Narasimhan, and Y. Cao, 
"React: Synergizing Reasoning and Acting in Language Models," 
\textit{arXiv preprint arXiv:2210.03629}, 2022.

\bibitem{du2023power}
Y. Du, S. C. Liew, K. Chen, and Y. Shao, 
"The Power of Large Language Models for Wireless Communication System Development: A Case Study on FPGA Platforms," 
\textit{arXiv preprint arXiv:2307.07319}, 2023.

\bibitem{liu2023verilogeval}
M. Liu, N. Pinckney, B. Khailany, and H. Ren, 
"VerilogEval: Evaluating Large Language Models for Verilog Code Generation," 
in \textit{2023 IEEE/ACM International Conference on Computer Aided Design (ICCAD)}, 2023, pp. 1--8.

\bibitem{fu2023llm4sechw}
W. Fu, K. Yang, R. G. Dutta, X. Guo, and G. Qu, 
"LLM4SecHW: Leveraging Domain-Specific Large Language Model for Hardware Debugging," 
\textit{Asian Hardware Oriented Security and Trust (AsianHOST)}, 2023.

\bibitem{thakur2023benchmarking}
S. Thakur, B. Ahmad, Z. Fan, H. Pearce, B. Tan, R. Karri, B. Dolan-Gavitt, and S. Garg, 
"Benchmarking Large Language Models for Automated Verilog RTL Code Generation," 
in \textit{2023 Design, Automation \& Test in Europe Conference \& Exhibition (DATE)}, 2023, doi: 10.23919/DATE56975.2023.10137086. [Online]. Available: \url{https://par.nsf.gov/biblio/10419705}.

\bibitem{xu2024genaipoweredmultiagentparadigmsmart}
H. Xu, J. Yuan, A. Zhou, G. Xu, W. Li, X. Ban, and X. Ye, 
"GenAI-powered Multi-Agent Paradigm for Smart Urban Mobility: Opportunities and Challenges for Integrating Large Language Models (LLMs) and Retrieval-Augmented Generation (RAG) with Intelligent Transportation Systems," 
\textit{arXiv preprint arXiv:2409.00494}, 2024.

\bibitem{zou2023wireless}
H. Zou, Q. Zhao, L. Bariah, M. Bennis, and M. Debbah, 
"Wireless Multi-Agent Generative AI: From Connected Intelligence to Collective Intelligence," 
\textit{arXiv preprint arXiv:2307.02757}, 2023.

\bibitem{islam2024mapcoder}
M. A. Islam, M. E. Ali, and M. R. Parvez, 
"MapCoder: Multi-Agent Code Generation for Competitive Problem Solving," 
\textit{arXiv preprint arXiv:2405.11403}, 2024.

\end{thebibliography}
\end{document}